\documentclass[lettersize,journal]{IEEEtran}
\usepackage{amsmath,amsfonts}
\usepackage{algorithmic}
\usepackage{algorithm}
\usepackage{array}
\usepackage[caption=false,font=normalsize,labelfont=sf,textfont=sf]{subfig}
\usepackage{textcomp}
\usepackage{stfloats}
\usepackage{url}
\usepackage{verbatim}
\usepackage{graphicx}
\usepackage{cite}
\usepackage{amssymb}
\usepackage{float}
\usepackage{times}
\usepackage{epsfig}
\usepackage{booktabs}
\usepackage{multirow}
\usepackage{setspace}
\usepackage{xcolor}
\usepackage{hyperref}

\hypersetup{ colorlinks,  citecolor=black,  linkcolor=black,  urlcolor=blue}

\begin{document}

\title{ClusVPR: Efficient Visual Place Recognition with Clustering-based Weighted Transformer}

\author{Yifan~Xu, Pourya~Shamsolmoali, and  Jie~Yang
\thanks{Y.~Xu, and J.~Yang are with the Institute of Image Processing and Pattern Recognition, Shanghai Jiao Tong University, Shanghai, China. Emails: \{yifanxu, jieyang\}@sjtu.edu.cn}
\thanks{P. Shamsolmoali is with the School of Communication and Electrical Engineering, East China Normal University, Shanghai, China. Email: pshams55@gmail.com} 
}

\markboth{}%
{Shell \MakeLowercase{\textit{et al.}}: A Sample Article Using IEEEtran.cls for IEEE Journals}


\maketitle

\begin{abstract}
Visual place recognition (VPR) is a highly challenging task that has a wide range of applications, including robot navigation and self-driving vehicles. VPR is particularly difficult due to the presence of duplicate regions and the lack of attention to small objects in complex scenes, resulting in recognition deviations. In this paper, we present ClusVPR, a novel approach that tackles the specific issues of redundant information in duplicate regions and representations of small objects. Different from existing methods that rely on Convolutional Neural Networks (CNNs) for feature map generation, ClusVPR introduces a unique paradigm called Clustering-based Weighted Transformer Network (CWTNet). CWTNet leverages the power of clustering-based weighted feature maps and integrates global dependencies to effectively address visual deviations encountered in large-scale VPR problems. We also introduce the optimized-VLAD (OptLAD) layer that significantly reduces the number of parameters and enhances model efficiency. This layer is specifically designed to aggregate the information obtained from scale-wise image patches. Additionally, our pyramid self-supervised strategy focuses on extracting representative and diverse information from scale-wise image patches instead of entire images, which is crucial for capturing representative and diverse information in VPR. Extensive experiments on four VPR datasets show our model's superior performance compared to existing models while being less complex.
\end{abstract}


\section{Introduction}
\begin{figure}[t]
\centering
   \includegraphics[width=0.98\linewidth]{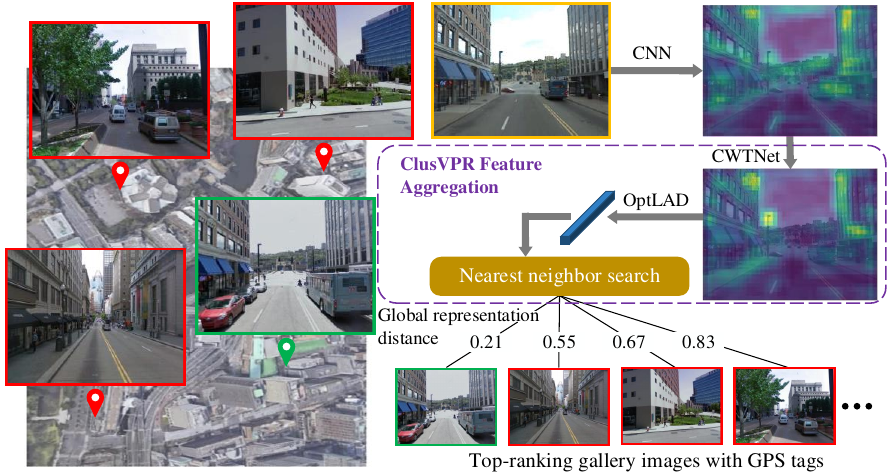}
   \caption{An illustration of the visual place recognition task using our model.}
\label{fig:taser_fig}
\end{figure}

\IEEEPARstart{V}{isual} Place Recognition (VPR), also known as image Geo-localization, is a technique that aims to determine the location of query images from unknown locations by comparing them against a database of reference images from known locations  \cite{arandjelovic_2018_netvlad, Fan_2022_TMM}. VPR plays a significant role in various AI applications, including robot navigation \cite{Khaliq_2020_Robotics} and self-driving vehicle \cite{HANE_2017_ICV}.
Recent advancements in deep CNNs \cite{He_2016_CVPR, Hsu_2018_TMM} and Vision Transformers (ViTs) \cite{Alexander_2021_ICLR, Lin_2021_TMM} have significantly improved the performance of VPR methods by extracting more discriminative and comprehensive features from images \cite{Ge_2020_ECCV, yang_2021_NIPS, Hausler_2021_CVPR, leyva2023data}.
In this paper, we focus on the VPR problem as an example of an image retrieval task. Our objective is to determine the location of an image by comparing it with geo-tagged images of a reference database.
The illustration of our VPR architecture is shown in Fig~\ref{fig:taser_fig}.

The main challenge in VPR is to generate discriminative representations from images with varying perspectives and appearances, such as environmental variations (e.g., seasons, illumination changes, dynamic occlusions). CNNs with NetVLAD \cite{Arandjelovic_2016_CVPR} are commonly used in VPR \cite{Ge_2020_ECCV, Hausler_2021_CVPR, ali2023global} due to their ability to extract comprehensive and discriminative features from images. However, their effectiveness is limited in complex scenes with visual distractions, such as occlusions and dynamic objects, because of the locality assumption of CNNs. In such cases, the extracted features may not be able to capture the important information for VPR, leading to inaccurate localization.
To tackle this issue, recent studies \cite{Alaaeldin_2021_arXiv, yang_2021_NIPS, Zhu_2023_arXiv} have explored the use of ViTs that have shown promising results in various vision tasks. However, ViTs face specific challenges when applied to VPR tasks.

\noindent {\bf First}, ViTs simply process the entire image as a sequence of image tokens and cannot explicitly consider the importance of individual image regions. This can be problematic in VPR tasks where certain regions, such as duplicate regions (regions with similar visual content) or small objects, can have a significant impact on recognition accuracy. Ignoring these regions can lead to deviations and inaccurate localization.
{\bf Second}, ViTs lack certain inductive biases that are present in CNNs, such as locality and translation equivariance. These biases are crucial for effectively learning from image data and capturing spatial relationships between image features.
{\bf Third}, VPR datasets often have noisy GPS labels, which can introduce uncertainties and inconsistencies in the training process. Additionally, relying on weakly-supervised learning methods \cite{Ge_2020_ECCV, Hausler_2021_CVPR} limits the availability of precise annotations, which can negatively impact the performance and accuracy of ViTs in VPR tasks.

Based on the above observations, we introduce a new model called ClusVPR. Our ClusVPR consists of CWTNet and OptLAD layers. By leveraging the self-attention property, the CWTNet enhances the model's ability to capture global context information. Additionally, we address the issue of the unequal importance of image regions in VPR by calculating weights for image tokens using the k-nearest neighbor (KNN) clustering algorithm. This allows us to assign different weights to image tokens based on their relevance and significance and effectively minimize visual deviations that commonly arise in complex VPR tasks. To improve computational efficiency, we incorporate sparse attention \cite{Child_2019_arXiv}, into the CWTNet. Sparse attention allows us to focus on the most relevant image tokens and minimize ineffective computations.
Furthermore, compared to the NetVLAD, OptLAD layer in ClusVPR plays a key role in optimizing the model's efficiency. It achieves this by splitting high-dimensional representation vectors into uniform groups and effectively reducing the dimension of the global descriptor.

Recent works \cite{Liu_2019_ICCV, Hausler_2021_CVPR} have addressed the challenge of noisy GPS labels in VPR by adopting approaches that select the most similar image to the query image in the representation space as the top-ranked positive (ground-truth) image.
These works are trained by using the triplet loss \cite{Schroff_2015_CVPR} to generate feature representations for the query image that are similar to the top-ranked positive image. However, the effectiveness of these methods is often limited due to the lack of information in the top-ranked positive images and the high appearance variability in large-scale VPR datasets, leading to instability and a lack of robustness. One potential solution to address this issue is to use high-ranked positive images during the training process. These images, although less similar to the query images, can provide valuable and diverse information that is often overlooked.
To address this, we propose a pyramid self-supervised strategy that allows us to extract accurate information from scale-wise image patches, taking advantage of the diverse information provided by high-ranked positive images. 
More precisely, we split and merge the input images into scale-wise patches and generate local representations. We compare the local representations between the input images to compute scale-wise similarity scores. These scores are then used to refine the pyramid self-supervision, enabling the model to learn and capture the variations and details present in the image patches at different scales.
In this strategy, the training process is divided into multiple generations to gradually improve the accuracy of the scale-wise similarity scores. 

The major innovations and contributions of this paper are as follows:

\begin{itemize}
\item We introduce ClusVPR, a novel approach that aggregates clustering-based weighted feature maps into a compact and discriminative representation. This approach enhances global context inference and improves the processing of unequally important image regions. Additionally, our OptLAD layer reduces model complexity and enhances robustness to variations and noise.
\item{A pyramid self-supervised strategy is designed to extract more representative and diverse information from scale-wise image patches. This approach avoids learning from entire images and instead focuses on extracting information from each scale level.}
\end{itemize}

The remainder of the paper is organized as follows: we first discuss the related works in Section \ref{sec:relatedwork}. Then, we detail the proposed ClusVPR and pyramid selfsupervised strategies in Section \ref{sec:methods}. In Section \ref{sec:experiments}, we provide the implementation details and experimental results. Finally, in Section \ref{sec:conclusions} we conclude the paper.

\section{Related work}
\label{sec:relatedwork}
\subsection{Recent progress on visual place recognition (VPR)}
VPR aims to establish a mapping between an image and its corresponding spatial location. In VPR, the goal is to find the geographical location of a query image, and the predicted coordinates are considered correct if they are close to the ground truth position. Early VPR approaches \cite{Cummins_2008_IJRR, Mur_2017_TR} relied on classical techniques like SIFT \cite{Lowe_1999_ICCV} and SURF \cite{Bay_2006_ECCV} for detecting keypoints. However, these handcrafted features are not suitable for dealing with the large variations in appearance that are encountered in complex scenes. In recent years, CNN-based methods have demonstrated superior performance in VPR \cite{Arandjelovic_2016_CVPR, Hausler_2021_CVPR, Zhu_2023_arXiv}. Among deep learning representation methods, NetVLAD \cite{arandjelovic_2018_netvlad} has emerged as a highly effective technique for VPR tasks. NetVLAD is a differentiable implementation of VLAD that is trained end-to-end with a CNN backbone for direct place recognition. It has been widely adopted in various works \cite{Ge_2020_ECCV, Hausler_2021_CVPR, Xu_2023_WACV, ali2023global}. Task-specific patch-level features have also been explored for VPR \cite{cao2020unifying, Ge_2020_ECCV}. 

The Patch-NetVLAD \cite{Hausler_2021_CVPR} introduces the use of the NetVLAD to extract descriptors from predefined image patches. Indeed, patch-level descriptors typically referred to local descriptors, encoding content from local patches. There are several researches on developing more compact descriptors, either through dimensionality reduction techniques \cite{zhu2018attention, cao2020unifying} or by replacing NetVLAD with lighter pooling layers such as GeM \cite{radenovic2018fine} and R-MAC \cite{gordo2017end}. Recently, Berton et al. \cite{berton2022deep} introduced a benchmark for VPR, providing a standardized framework to evaluate various global-retrieval-based methods. This benchmark allows for a fair comparison between different approaches in the field. In \cite{leyva2023data}, a generalized contrastive Loss (GCL) is developed for training networks using graded similarity labels and \cite{shen2023structvpr} considers knowledge distillation to build an efficient VPR method. 

To summarize, the existing models that use the entire image for training often lead to inaccurate global representation learning. To address this issue, our proposed method leverages a pyramid self-supervised strategy, which involves using scale-wise image patches instead of the entire image. This approach enables the model to capture diverse and representative information during the training process.

\begin{figure*}[t]
\begin{center}
  \includegraphics[width=0.98\linewidth]{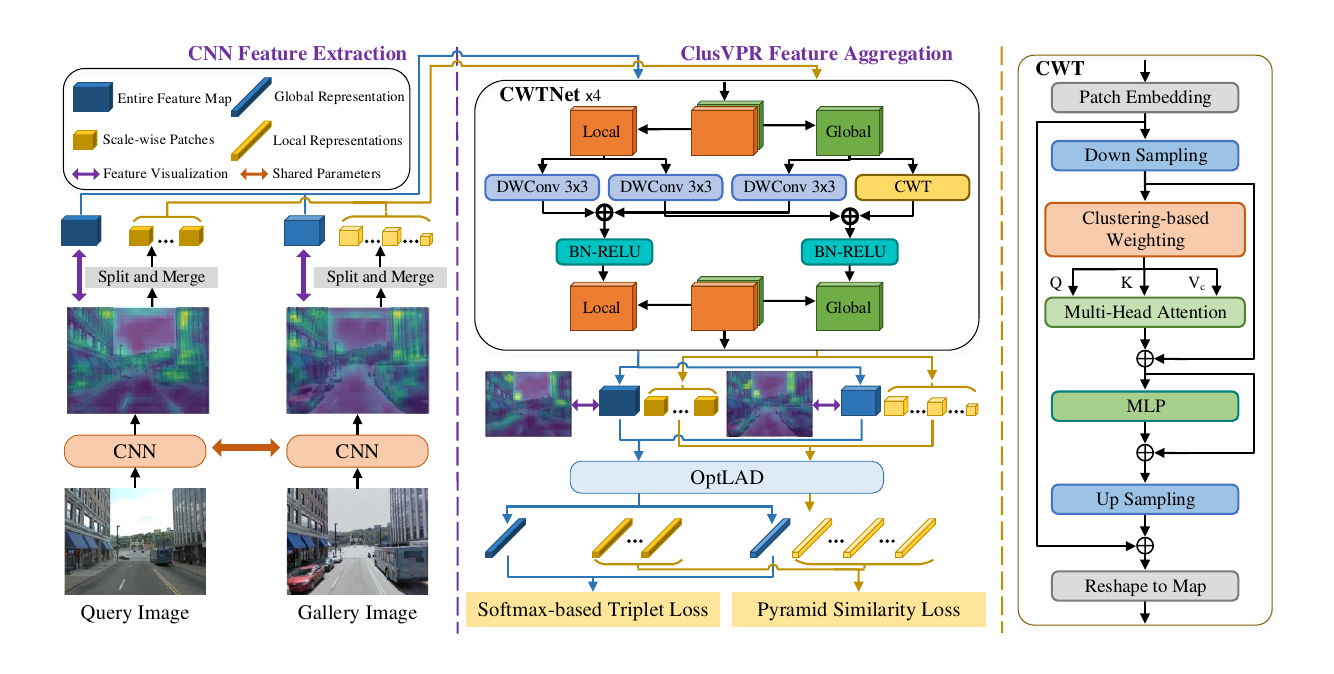}
\end{center}
\vspace{-5pt}
   \caption{The pipeline of the training architecture. Our model consists of a CNN backbone and the proposed ClusVPR.}
\label{fig:frame}
\end{figure*}

\subsection{Transformer-based Approaches}
The Transformer \cite{Vaswani_2017_NIPS} was initially proposed for Natural Language Processing tasks and later introduced to vision tasks as the ViT \cite{Alexander_2021_ICLR}. In ViT, each image patch is treated as a token, allowing the model to process images using a sequence-based approach. The vanilla ViT requires large-scale training datasets, such as ImageNet \cite{gordo2017end}, to achieve comparable results to CNNs \cite{He_2016_CVPR}. DeiT \cite{Touvron_2021_ICML}, on the other hand, introduces a data-efficient training strategy for ViT that outperforms CNNs on the average-scale datasets.
The vanilla ViT has been adopted in the recent VPR benchmarks, demonstrating competitive performance on global retrieval tasks. \cite{Alaaeldin_2021_arXiv}  applied ViT to image retrieval tasks by utilizing the [class] token from the final layer as a global feature. \cite{Wang_2022_CVPR} takes a different approach by incorporating multiple transformer layers while still relying on a CNN-based backbone feature extractor. DELG \cite{cao2020unifying} is a framework that extracts both local and global features. However, to improve the representation of features, HAF \cite{Yan_2021_ICASSP} is introduced which consist of a hierarchical attention network. DASGIL \cite{hu2020dasgil} presents a multi-task architecture with a shared encoder to generate global representation.
Building upon this, \cite{paolicelli2022learning} focuses on filtering semantic information using an attention mechanism.
These approaches while promising, but they often overlook the specific challenges faced in VPR tasks, such as the varying importance of image regions due to duplicate regions and ignored small objects. In our ClusVPR, we tackle this challenge by introducing a novel approach that computes weights for image tokens using the k-nearest neighbor (KNN). This enables us to reduce visual deviations and enhance the accuracy of VPR.

\section{Proposed Model}
\label{sec:methods}
This section presents a detailed description of our ClusVPR model. The training architecture is shown in Fig~\ref{fig:frame}.

\subsection{CWTNet of the ClusVPR}
To address the issue of deviations in VPR tasks caused by duplicate regions and overlooked small objects, we introduce CWTNet, which computes clustering-based weighted feature maps. CWTNet divides channels into two parallel branches, namely the local and global branches.
The local branch uses depth-wise convolution layers to extract local features and provides positional information for our Clustering-based Weighted Transformer (CWT), because $3 \times 3$ convolutions can provide sufficient positional information \cite{wu_2021_ICCV}, eliminating the need for positional embedding as used in ViTs. 
The global branch uses the proposed CWT to obtain global context. The local and global features are then fused through feature concatenation to generate the final output. The structure of the CWTNet is shown in Fig~\ref{fig:frame}.

ViTs \cite{Alexander_2021_ICLR, Touvron_2021_ICML} typically take a sequence of 1D token embeddings as input. Given a feature map $\textbf{m} \in {\bf R}^{C\times H\times W}$, we first split it into a number of individual feature patches $\textbf{m}_p\in{\bf R}^{N \times P^2 \cdot C}$, followed by their subsequent down-sampling into corresponding feature vectors $\textbf{x}_p\in{\bf R}^{N \times C}$. Here, $(H,W)$ represents the resolution of the feature map, $C$ denotes the number of channels, $P$ is the down-sampling rate, and $N = HW/P^2$ represents the total number of token embeddings. Subsequently, $\textbf{x}_p$ is used for the input token embeddings. 
We introduce a clustering-based weighted module that uses the KNN to calculate the weights for the input token embeddings $\textbf{x}_p$.
Given a set of token embeddings $\textbf{x}_p$, the clustering density $\rho_i$ of each token embedding $x_i$ is computed through its $k_n$-nearest neighbors:
\begin{equation}
    d_i = \sum_{x_j \in \text{KNN}(x_i)} {\|x_i - x_j\|}^2_2,
    \label{Eq.1}
\end{equation}
\vspace{-5pt}
\begin{equation}
    \rho_i = \text{exp} (-\frac{d_i - d_{min}}{k(d_{max} - d_{min})}),
    \label{Eq.2}
\end{equation}

\noindent where $x_i$ and $x_j$ are the corresponding token embeddings, and $\text{KNN}(x_i)$ represents the $k_n$-nearest neighbors of $x_i$. $d_i$ is the sum of distances between $x_i$ and its $k_n$-nearest neighbors. The weight of each token embedding $x_i$ is computed by
\begin{equation}
    w_i = \frac{1 - \rho_i}{\sum_{j = 1}^N (1 - \rho_j)},
    \label{Eq.3}
\end{equation}
\begin{equation}
    w_i = \frac{w_i - w_{min}}{w_{max} - w_{min}},
    \label{Eq.4}
\end{equation}

\noindent where $w_i$ denotes the potential value of each token embedding. We use ViT \cite{Alexander_2021_ICLR} as the basis for the CWT, but we remove the Layer Normalization (LN) to enable our clustering-based weighted module to operate more effectively.

In the clustering-based weighted multi-head self-attention (CMSA), we use the clustering-based weights to compute the self-attention head. Given the token embeddings $\textbf{x}_p$ as the input sequence of the CWT, a single self-attention head can be represented by
\begin{equation}
    \textbf{Q} = \textbf{x}_p \textbf{W}_q, \ \textbf{K} = \textbf{x}_p \textbf{W}_k, \ \textbf{V} = \textbf{x}_p \textbf{W}_v, 
    \label{Eq.5}
\end{equation}
\begin{equation}
    \textbf{A} = \text{softmax}(\frac{\textbf{Q} \textbf{K}^T}{\sqrt{D_k}})
    ((\lambda_c\textbf{I}_v + \textbf{W}_c)\textbf{V}).
    \label{Eq.6}
\end{equation}
here $\textbf{W}_q$, $\textbf{W}_k$, $\textbf{W}_v$ are linear projection matrices, and $\textbf{Q}$, $\textbf{K}$, $\textbf{V}$ represent the query, key, and value matrices. $D_k$ denotes the channel quantity of the token embeddings, and $\textbf{I}_v$ is an identity matrix with the same rank as $\textbf{V}$. $\textbf{W}_c$ is a diagonal matrix whose main diagonal elements are the clustering-based weights $w_i$. Finally, $\lambda_c$ is a scaling factor, which we set to $0.5$. Thus, our CWT is formulated as follows:
\begin{equation}
    \textbf{z}_p = \text{CMSA}(\textbf{x}_p) + \textbf{x}_p, \ 
    \textbf{x}_p = \text{DS}(\textbf{m}_p),
    \label{Eq.7}
\end{equation}
\begin{equation}
    \textbf{z}'_p = \text{MLP}(\textbf{z}_p) + \textbf{z}_p,
    \label{Eq.8}
\end{equation}
\begin{equation}
    \textbf{m}_{\text{out}} = \text{US}(\textbf{z}'_p) + \textbf{m}_p.
    \label{Eq.9}
\end{equation}
where the CWT consists of a down-sampling layer (DS), a CMSA module, a multi-layer perceptron (MLP), and an up-sampling layer (US). The DS is implemented using average pooling. The MLP comprises a GELU \cite{Hendrycks_2016_arXiv} activation layer and two linear layers. The US is implemented using depthwise separable transposed convolution, and the output sequence $\textbf{m}_{\text{out}}$ is reshaped into a clustering-based weighted feature map $\textbf{m}_c \in {\bf R}^{C\times H\times W}$.
\subsection{OptLAD of the ClusVPR}
\label{OptLAD}
Considering that the high dimensionality of global descriptors obtained by the standard NetVLAD is unsuitable for large-scale VPR tasks, a PCA layer with whitening \cite{Herv_2012_ECCV} is adopted to reduce the dimensionality to a suitable value $N'$. However, this approach requires a large number of parameters. 
For example, for aggregating a feature map with $1024$ channels through $128$ clusters, using a PCA layer with a $4096$-dimension output vector would require about $537$M parameters. This makes it difficult to deploy the model on resource-constrained mobile devices. To address this issue, we reduce the dimension of global descriptors, thereby reducing the number of parameters required by the PCA layer. This is achieved by splitting the input vector into uniform clusters of relatively low-dimensional vectors prior to executing the VLAD operation. Furthermore, a group weight $\beta_g$ calculated by GeM \cite{radenovic2018fine} is introduced to improve the non-linearity ability of the OptLAD.
\begin{figure}
\begin{center}
  \includegraphics[width=0.98\linewidth]{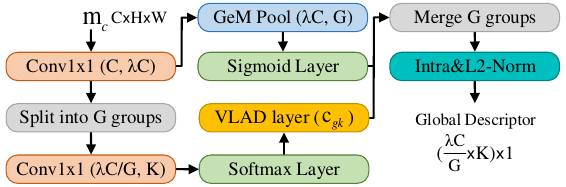}
\end{center}
\vspace{-5pt}
   \caption{The OptLAD structure. The primary parameters for each layer are indicated in brackets.}
\label{fig:vlad_layer}
\end{figure}

As Fig~\ref{fig:vlad_layer} illustrates the details, the input feature map $\textbf{m}_c\in{\bf R}^{C\times H\times W}$ has $D$ $C$-dimensional pixel-level descriptors ($D = HW$). The pixel-level descriptors are first expanded by $\lambda$ via a $1\times1$ convolution. Then, the input descriptors $\{\hat{x}_{i}\}, i=1,\cdots,D$ are split into G uniform groups. The low-dimensional descriptors are shown by $\{\hat{x}_{gi}\}, g=1,\cdots,G$. Given a set of cluster centers for each group $\{c_{gk}\}, k=1,\cdots, K$, the output of the global descriptor $V$ is represented as a $\frac{\lambda CK}{G}$-dimension vector. To better understand the output of $V$, we can also express it as a $\frac{\lambda C}{G}\times K$ matrix. Therefore, the element (j, k) of $V$ can be computed by
\begin{equation}
    V(j,k,g) = \sum_{i=1}^{D} \alpha_{k}(\hat{x}_{gi}) (\hat{x}_{gi}(j) - c_{gk}(j)),
    \label{Eq.10}
\end{equation}
\vspace{-5pt}
\begin{equation}
    V(j,k) = \sum_{g=1}^{G} \beta_g(\hat{x}) V(j,k,g),
    \label{Eq.10_2}
\end{equation}
where $\hat{x}_{gi}(j)$ is the $j$-th dimension of the low-dimensional descriptor $\hat{x}_{gi}$. $c_{gk}(j)$ is the $j$-th dimension of the cluster center $c_{gk}$. $\alpha_{k}$ is used to find the proximity between the lower-dimensional descriptor $\hat{x}_{gi}$ and the cluster center $c_{gk}$. The $\beta_g$ is used to merge the output of each group.
\begin{equation}
    \alpha_{k}(\hat{x}_{gi}) = \frac{\text{exp}(w_{gk}^T \hat{x}_{gi}+ b_{gk})}
    {\sum_{k'=1}^K \text{exp}(w_{gk'}^T \hat{x}_{gi}+ b_{gk'})},
    \label{Eq.11}
\end{equation}
\begin{equation}
    \beta_g(\hat{x}) = \sigma \left(\frac{1}{D} \sum_{i=1}^{D} \left(\frac{1}{D'} \sum_{j=1}^{D'} \hat{x}_{gi}(j)^{p_i}\right)^{\frac{1}{p_i}}\right).
    \label{Eq.12}
\end{equation}
where $D'=\frac{\lambda C}{G}$ is the dimension of $\hat{x}_{gi}$, and $\sigma(\cdot)$ indicates the sigmoid function. The first weight $\alpha_{k}(\hat{x}_{gi})$, represents the distance between each pixel-level descriptor $\hat{x}_{gi}$ and the cluster center $c_{gk}$, while the second weight $\beta_g(\hat{x})$, is a scaling coefficient across groups.

The matrix $V$ is obtained by Intra-Normalization \cite{Arandjelovic_2013_CVPR} and then re-converted into a $\frac{\lambda CK}{G}$-dimensional vector, which is L2-normalized to obtain a compact global descriptor. 
Finally, we apply a PCA layer to reduce the dimension of the global descriptor to a proper value $N'$. Compared to the standard NetVLAD, the OptLAD with PCA requires approximately $\frac{\lambda}{G}$ times fewer parameters.

\subsection{Pyramid self-supervised strategy}
The VPR datasets \cite{Torii_2015_CVPR, Arandjelovic_2016_CVPR} lack fine-grained annotations and only provide GPS locations. Recent works \cite{Liu_2019_ICCV, Hausler_2021_CVPR} have adopted a simplified evaluation protocol, considering only the top-ranked positive image as the ground truth. During training, these models use the standard triplet loss \cite{Schroff_2015_CVPR}, which encourages the feature representation of a query image to be closer to its top-ranked positive image.
However, due to the limited information in top-ranked positive images, this approach lacks robustness. Instead, we use high-ranked positives to gather comprehensive information from scale-wise image patches.

\begin{figure}[t]
\centering
  \includegraphics[width=0.98\linewidth]{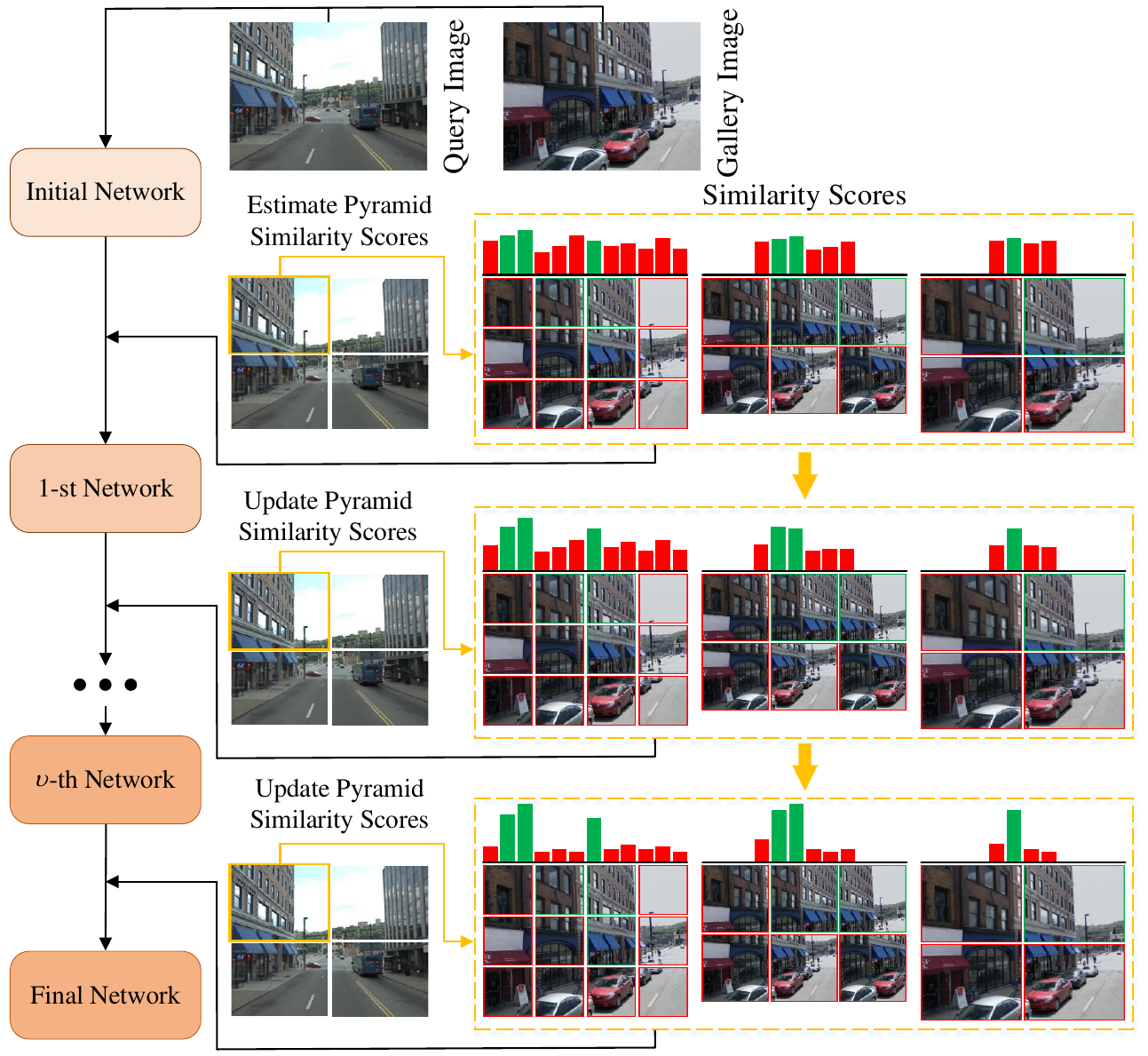}
  \vspace{-5pt}
   \caption{The proposed pyramid self-supervised strategy and how scale-wise patches are split and merged, where the green borders indicate high similarity with the query patch while the red borders denote low similarity. The pyramid similarity scores are progressively updated during training.}
\label{fig:multi_scale_scores}
\end{figure}

In particular, for a query image $q$ and a high-ranked positive image $p^l$, the feature maps $\{\textbf{m}_{q}^{\theta_0}, \textbf{m}_{p^l}^{\theta_0}\}$ are extracted from the CNN backbone, where $\theta_0$ represents the network's initial parameters. To obtain scale-wise patches, we carry out feature maps $\{\textbf{m}_{q}^{\theta_0}, \textbf{m}_{p^l}^{\theta_0}\}$ splitting and merging operations as evidently depicted in Fig~\ref{fig:multi_scale_scores}. Then, we feed the resulting feature maps $\{\textbf{m}_{q_1}^{\theta_0}, \cdots, \textbf{m}_{q_{4}}^{\theta_0}, \textbf{m}_{p^l_1}^{\theta_0}, \cdots, \textbf{m}_{p^l_{22}}^{\theta_0}\}$ separately to the ClusVPR to generate local representations $\{f_{q_1}^{\theta_0}, \cdots, f_{q_{4}}^{\theta_0},$ $ f_{p^l_1}^{\theta_0}, \cdots, f_{p^l_{22}}^{\theta_0}\}$. To compute similarity between $q$ and $p^l$, we calculate pyramid similarity scores by
\begin{equation}
\begin{aligned}
    \mathcal{S}_{\theta_0}(\tau_0)=\text{softmax}([\langle f_{q_1}^{\theta_0},f_{p^l_1}^{\theta_0}\rangle /\tau_0, \cdots, \langle f_{q_1}^{\theta_0}, f_{p^l_{22}}^{\theta_0}\rangle /\tau_0, \\ \cdots, \langle f_{q_{4}}^{\theta_0}, f_{p^l_1}^{\theta_0}\rangle /\tau_0, \cdots, \langle f_{q_{4}}^{\theta_0},f_{p^l_{22}}^{\theta_0}\rangle /\tau_0]),
    \label{Eq.13}
\end{aligned}
\end{equation}
where $\tau_0$ is a hyperparameter that influences the level of smoothness in the pyramid similarity scores $\mathcal{S}_{\theta_0}$, and $\langle\cdot,\cdot\rangle$ represents the inner product operation.

We enhance pyramid similarity scores through $\omega$ generations, as illustrated in Fig~\ref{fig:multi_scale_scores}. The $(\upsilon-1)$-th generation's pyramid similarity $\mathcal{S}_{\theta{\upsilon-1}}$ supervises the $\upsilon$-th network using Kullback-Leibler (KL) divergence. The pyramid similarity loss for a high-ranked positive image $p^l$ is written as:
\begin{equation}
    \mathcal{L}_{s}^{\theta_{\upsilon}}(q,p^l)=\ell_{kl}(\mathcal{S}_{\theta_{\upsilon}}(1),\mathcal{S}_{\theta_{\upsilon-1}}(\tau_{\upsilon-1})).
    \label{Eq.14}
\end{equation}
where $\theta_{\upsilon}, \upsilon=1,\cdots$ and $\omega$ are the parameters associated with the $\upsilon$-th generation network. $\ell_{kl}$ represents the KL divergence, which can be written as $\ell_{kl}(y,\hat{y}) =\sum_{i}\hat{y}(i)\log(\hat{y}(i) / y(i))$.
The target similarity vector $\mathcal{S}_{\theta_{\upsilon-1}}$ is generated using the softmax, with a temperature coefficient $\tau_{\upsilon-1}$ to control its smoothness. A larger temperature coefficient leads to a more uniformly distributed similarity vector, meaning that the pyramid similarity loss focuses on more matching pairs. So, in the early generations, $\tau_{\upsilon-1}$ is set to a larger value.

\begin{table*}[t]
    \centering
    \small{
    \caption{The precision and complexity comparison of our model on four VPR benchmarks.}
    \label{tab.geo_recalls}
    \setlength{\tabcolsep}{1.4mm}{
    \begin{tabular}{ l| c  c  |c  c  | c  c  | c  c  | c  c  | c  c  | c  c }
        \toprule[0.5pt]
        \multirow{2}{4em}{\textbf{Method}}& \multicolumn{2}{c|}{\textbf{MSLS val}} & \multicolumn{2}{c|}{\textbf{MSLS chall}} & \multicolumn{2}{c|}{\textbf{Pitts30k}} & \multicolumn{2}{c|}{\textbf{Pitts250k}} & \multicolumn{2}{c|}{\textbf{Tokyo 24/7}} & \multicolumn{2}{c|}{\textbf{TokyoTM}} & \multicolumn{2}{c}{\textbf{Model Complexity}} \\ \cmidrule(r){2-15}
    		
        & R@1 & R@5& R@1 & R@5  & R@1 & R@5  & R@1 & R@5  & R@1 & R@5  & R@1 & R@5  & Params(M) & FLOPs(G)\\
        \midrule[0.5pt]
        NetVLAD \cite{Arandjelovic_2016_CVPR} &70.1 &80.8&41.4 &53.7& 85.6 & 92.9  & 86.0 & 93.2  & 73.3 & 82.9  & 93.9 & 96.8  & 148.97 & 94.35 \\ [-0.5ex]

        SARE \cite{Liu_2019_ICCV} &76.4 &81.6&44.3 &55.8  &87.2 &93.8& 89.0 & 95.5  & 79.7 & 86.7  & 94.5 & 96.7  & 148.97 & 94.35 \\ [-0.5ex]

        SFRS \cite{Ge_2020_ECCV} &77.2 &82.5&45.8 &56.1& 89.4 & 94.7  & 90.7 & 96.4  & 85.4 & 91.1  & 95.8 & 98.2  & 148.97 & 94.35 \\ [-0.5ex]


       HAF \cite{Yan_2021_ICASSP} &78.5 & 85.9& 47.3 & 56.3  &86.6 & 93.7& 89.4 & 95.8  & 78.2 & 84.4  & 94.8 & 97.8  & 158.87 & 1791.27 \\ [-0.5ex]

        Patch-NetVLAD \cite{Hausler_2021_CVPR} &79.5 &86.2&48.1 & 57.4& 88.7 & 94.5  & 89.8 & 95.9  & 86.0 & 88.6  & 95.2 & 97.8 & 148.97 & 94.22 \\ [-0.5ex]

        GPM \cite{ali2023global} &80.3 &86.8&48.5 &\bf{58.6}& 89.2 & 94.6  & 91.2 & 96.7  & 85.4 & 89.9  & 96.3 & 98.2 & 151.34 & 95.24 \\ [-0.5ex]

        GCL-NetVLAD \cite{leyva2023data} &78.0 &85.1&47.6 &57.4& 88.9 &93.8  & 91.5 & \bf{97.1}  & 85.7 & 90.2  & 96.2 & 98.0 & 148.97 & 94.35 \\ [-0.5ex]
        

       \midrule
        ClusVPR &\bf{82.7} & \bf{88.5} & \bf{49.4} & 58.2& \textbf{90.8} & \bf{95.2}  & \textbf{92.4} & {96.9}  & \textbf{87.5} & \textbf{91.5}  & \textbf{96.8} & \textbf{98.7}  & \textbf{53.12} & \bf{92.14} \\ [-0.5ex]   
        \midrule
    \end{tabular}}}
\end{table*}

Our training is supervised using a triplet loss, similar to \cite{Yan_2021_ICASSP, Hausler_2021_CVPR}. Specifically, each triplet consists of a query image $q$, a corresponding top-ranked positive image $p^*$, and a set of high-ranked negative images ($ n_i|_{i=1}^M$), where the rank of positive or negative images is determined by the Euclidean distance between their global representations. However, the standard triplet loss \cite{Schroff_2015_CVPR} lacks robustness and relies on the proper selection of negative images. Hence, a softmax-based triplet loss is used to notably increase the distinction among the positive pair and a set of negative pairs, as given below,
\begin{equation}
    \mathcal{L}_{t}^{\theta}(q,p^*,n) = \sum_{i=1}^{M}-\log\frac{\exp \langle f^{\theta}_q, f^{\theta}_{p^*} \rangle}{\exp \langle f^{\theta}_q, f^{\theta}_{p^*}\rangle + \exp \langle f^{\theta}_q, f^{\theta}_{n_i}\rangle},
    \label{Eq.16}
\end{equation}
where $\theta$ is the network's parameter, and $\epsilon$ is the margin between positive and negative pairs. The $p^*$ is chosen as the top-ranked images of the gallery that located within 10 meters from the $q$, while $\{n_i |_{i=1}^M\}$ represents a set of randomly chosen gallery images from the top 500, located at a distance of 25 meters away from $q$.

Therefore, our model is supervised by the pyramid similarity loss in conjunction with the softmax-based triplet loss. Given $K$ high-ranked positive images $\{p^k |_{k=1}^K\}$, the total loss can be computed by
\begin{equation}
    \mathcal{L}_{total}^{\theta_\upsilon} = \mathcal{L}_{t}^{\theta_\upsilon}(q,p^*,n) + \lambda_s \sum_{k=1}^K \mathcal{L}_{s}^{\theta_\upsilon}(q,p^k).
    \label{Eq.17}
\end{equation}
where $\lambda_s$ is the loss scaling coefficient.

\section{Experiments}
\label{sec:experiments}
This section evaluates the efficacy of our model on both VPR and image retrieval datasets compared to several state-of-the-art models \cite{Arandjelovic_2016_CVPR, Liu_2019_ICCV, Ge_2020_ECCV, Yan_2021_ICASSP, Hausler_2021_CVPR, ali2023global, leyva2023data}. Specifically, we show the details of datasets, implementations, and quantitative/qualitative results.

\subsection{Datasets}
Our model is evaluated on four VPR datasets -- the MSLS \cite{warburg2020mapillary}, Pitts30k/250k-test \cite{Torii_2013_CVPR}, Tokyo 24/7 \cite{Torii_2015_CVPR}, and TokyoTM-val \cite{Arandjelovic_2016_CVPR}. These datasets include GPS labels, and images have varying appearances and perspectives. Our model's performance is evaluated based on the recommended train-test settings of the datasets. To further demonstrate the generalization ability, we also assessed our model on three image retrieval datasets -- the Paris 6K \cite{Philbin_2008_CVPR}, Oxford 5K \cite{Philbin_2007_CVPR} and the Holidays \cite{Jegou_2008_ECCV}.

\subsection{Implementation details}
\subsubsection{Model settings}
The base ClusVPR contains four CWTNets for computing clustering-based weighted feature maps. For each CWTNet, the patch size on the feature map is set as $2 \times 2$, the down-sampling rate $P$ is $2$, and the number of nearest neighbors $k_n$ is set as $10$. For the OptLAD, the number of cluster centers $K$ is $64$, and we set $\lambda=2, G=8$. The dimension $N'$ of the global representation is $4096$. For a fair comparison, we use the VGG-16 backbone for feature extraction, similar to the other works.

\subsubsection{Training settings}
In our experiments, we train the model on Pitts30k-train and test it on Pitts30/250k-test, Tokyo 24/7-test, and TokyoTM-val, following standard VPR procedures. Additionally, we train the model on MSLS training set, and test it on MSLS validation and Challenge sets. The training consists of five generations, each comprising eight epochs.
We optimize the loss function using stochastic gradient descent (SGD) with a learning rate of 0.0001, weight decay of 0.001, and momentum of 0.9. We conducted a grid search to find the optimal hyperparameters. The loss scaling coefficient $\lambda_s$ is set to 0.55, and the temperature coefficient $\tau_{\upsilon}$ is set to 0.06.

\subsection{Comparison with the state-of-the-arts}
ClusVPR is compared with seven models--  NetVLAD \cite{Arandjelovic_2016_CVPR}, SARE \cite{Liu_2019_ICCV}, SFRS \cite{Ge_2020_ECCV}, HAF \cite{Yan_2021_ICASSP}, Patch-NetVLAD \cite{Hausler_2021_CVPR}, GPM \cite{ali2023global}, and GCL \cite{leyva2023data} with VGG-NetVLAD. We ensure a fair comparison by keeping the training set, cluster quantity, and dimension of global representations consistent with those used in other models. 

\subsubsection{VPR benchmarks}
Table~\ref{tab.geo_recalls} represents the results of our ClusVPR model in comparison with other baselines on the VPR benchmarks. The table reports precision-recall and model complexity metrics. The number of parameters (Params) includes parameters of the entire model. 
The FLOPs are computed based on an input image size (640, 480). The accuracy is measured using precision-recall. The top-$k$ recall, on the other hand, represents the percentage of query images that are correctly retrieved from the top-$k$ ranked gallery images.
The results presented in Table~\ref{tab.geo_recalls} demonstrate that our model outperforms other baselines on most benchmarks. Notably, our model surpasses the GPM, improving rank-1 recall by 2.4\%, 0.9\%, 1.6\%, 1.2\% 2.1\% and 0.5\%, respectively.
Moreover, our model generates only around one-third of the parameters produced by other models, indicating its efficient use of parameters.

\subsubsection{Image retrieval benchmarks}
Table~\ref{tab.imr_mAP} presents the mean Average Precision (mAP) results of our model compared to other state-of-the-art models on image retrieval benchmarks. 
In this experiment, all models are trained on Pitts30K and tested without any fine-tuning. For the Oxford 5K and Paris 6K, the models are evaluated on both full and cropped query images. The "Full" setting indicates that the entire image is taken as a query, while for "Crop", only the landmark region is used. For the Holidays dataset, the query images are used directly. 
\begin{table}
    \centering
    \small{
    \caption{Comparison of Mean Average Precision (mAP) on three image retrieval benchmarks.}
    \label{tab.imr_mAP}
    \setlength{\tabcolsep}{1.8mm}{
    \begin{tabular}{ l | c  c | c  c | c }
        \toprule[0.5pt]
        \multirow{2}{4em}{\textbf{Method}} & \multicolumn{2}{c|}{\textbf{Oxford 5K}} & \multicolumn{2}{c|}{\textbf{Paris 6K}} & \multirow{2}{4em}{\textbf{Holidays}} \\ \cmidrule(r){2-5}
	& full & crop & full & crop &\\
        \midrule[0.5pt]
	NetVLAD \cite{Arandjelovic_2016_CVPR} & 69.1 & 71.6 & 78.5 & 79.7 & {83.0} \\ [-0.6ex]
	SARE \cite{Liu_2019_ICCV} & 71.7 & 75.5 & 82.0 & 81.1 & 80.7 \\ [-0.6ex]
        SFRS \cite{Ge_2020_ECCV} & 73.9 & 76.7 & 82.5 & 82.4 & 80.5 \\[-0.6ex]
	HAF \cite{Yan_2021_ICASSP} & 71.9 & 75.8 & 82.3 & 81.5 & 81.4 \\ [-0.6ex]
	  Patch-NetVLAD \cite{Hausler_2021_CVPR} & 72.6 & 76.1 & 82.3 & 81.9 & 81.7 \\[-0.6ex]
        GPM \cite{ali2023global} & 73.3 & 76.4 & 82.8 & 82.3 & 82.3 \\ [-0.6ex]
        GCL-NetVLAD \cite{leyva2023data} & 73.5 & 76.7 & 82.7 & 82.5 & 82.5\\ [-0.6ex]
        \midrule
	ClusVPR  & \textbf{74.6} & \textbf{77.9} & \textbf{84.2} & \textbf{83.4} & \bf{83.8} \\[-0.6ex]
  
        \bottomrule[0.5pt]
    \end{tabular}}}
\end{table}
The results demonstrate that ClusVPR is effective in improving the performance of image retrieval. In particular, our model surpasses the performance of other models on all datasets, showing improvements of 1.1\%, 1.2\%, 1.5\%, 0.9\% and 1.3\% respectively, compared to GCL.

\subsection{Qualitative Evaluation}
\begin{figure*}[t]
\begin{center}
  \includegraphics[width=0.96\linewidth]{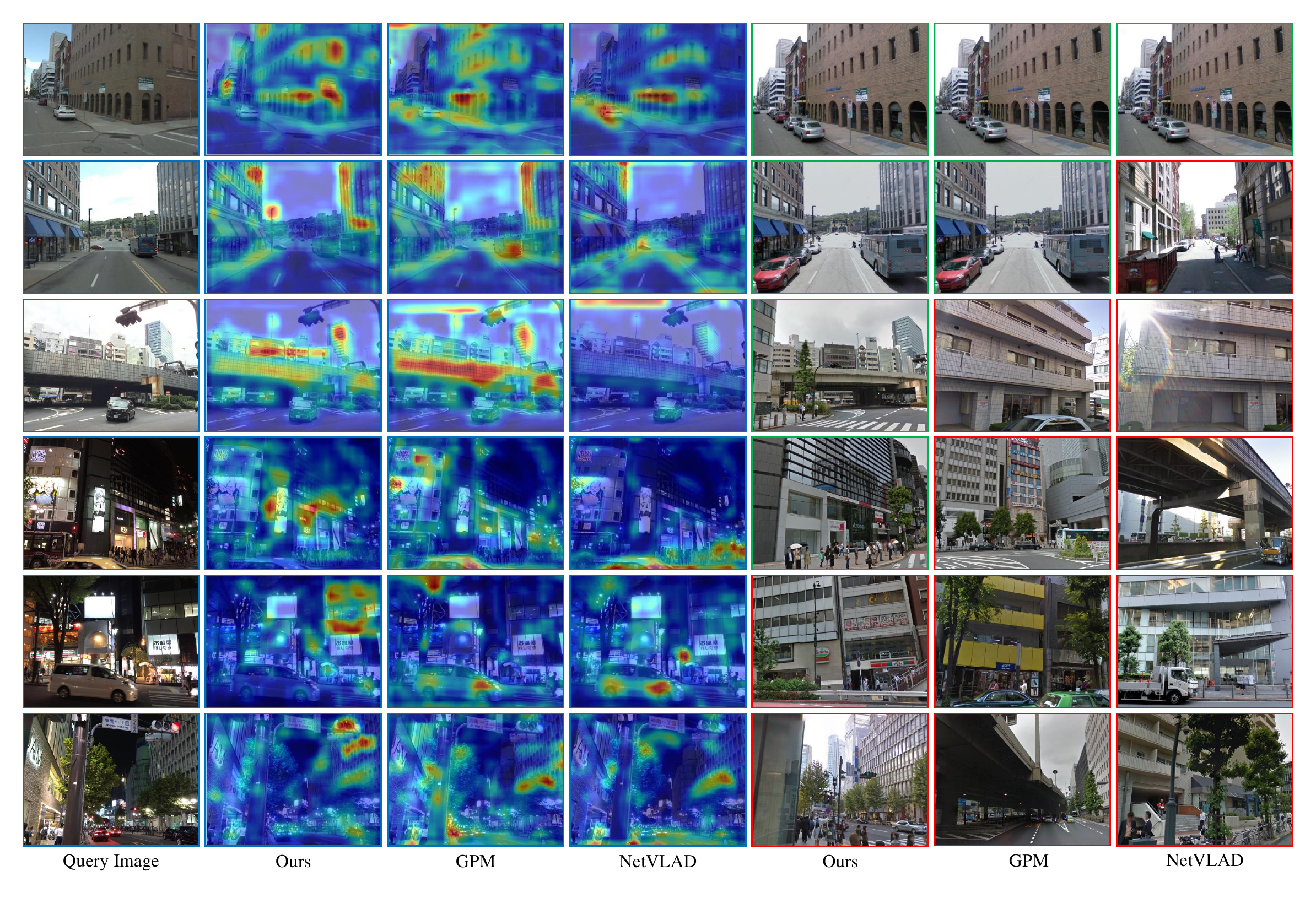}
\end{center}
   \caption{Visualization results of challenging cases. The attention maps overlaid on the images highlight the regions of interest identified by the models. Green and red borders around the top-1 retrieved images indicate successful or unsuccessful retrieval results, respectively.}
\label{fig:attention_compare}
\end{figure*}

The qualitative results of NetVLAD, GPM and our model on challenging cases with complex environments and varying lighting conditions can be seen in Fig~\ref{fig:attention_compare}. It's important to note that we used the feature maps before the VLAD operation for all test models and followed the approach detailed in \cite{Stylianou_2019_WACV} to produce the attention maps. 
In the first four cases, our model attends to discriminative regions such as signs and buildings, while the other two models wrongly focus on dynamic objects or obstacles such as pedestrians, lights, cars, and trees. In particular, in the third and fourth samples, where there are lighting issues and occlusions, our model successfully conducts matching based on discriminative landmarks while disregarding dynamic objects or barriers.
\begin{figure*}[t]
\begin{center}
   \includegraphics[width=0.96\linewidth]{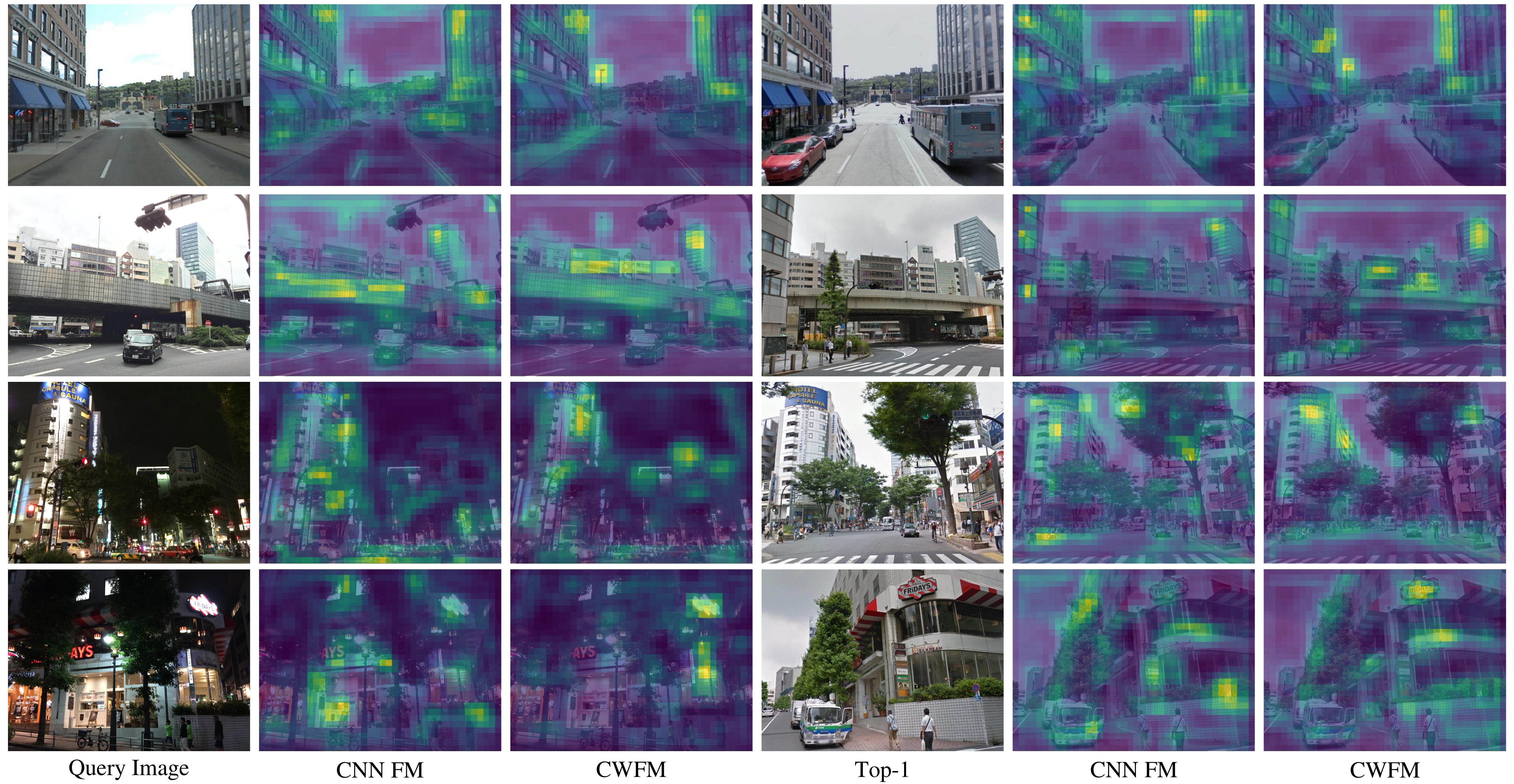}
\end{center}
   \caption{The visualization of CNN-based feature maps (CNN FMs) and clustering-based weighted feature maps (CWFMs) generated from the query images and their top-1 retrieved gallery images.} 
\label{fig:attention_feature_map}
\end{figure*}
However, the other two models seem to be focusing more on lights and cars, which are dynamic objects and obstacles that can appear or disappear in any gallery image. This could result in inaccurate retrieval results due to the distortions caused by such objects. In the last two challenging cases of Fig~\ref{fig:attention_compare}, where the query image contains complex scenes and lighting conditions, all models retrieved the wrong top-1 gallery images. However, our model still focuses on discriminative regions and retrieves a gallery image with a similar building structure.

To demonstrate the efficacy of the ClusVPR in addressing the unequal importance of image regions in VPR and improving global context inference, we conducted a comparison between CNN-based feature maps (the outputs of the CNN backbone) and clustering-based weighted feature maps (the outputs of the CWTNet). We selected two daytime and two nighttime query images from the Pitts30k-test and Tokyo 24/7 and show the results in Fig~\ref{fig:attention_feature_map}. The comparison demonstrates that the CWTNet of ClusVPR eliminates redundancy from duplicate regions and enhances discriminative information from small objects.

\subsection{Ablation experiments}
Several ablation experiments are conducted to evaluate the effectiveness of our ClusVPR and pyramid self-supervised strategy (PSS). The experiments aimed to evaluate the effectiveness of different components in our model and compare their performance. 
\begin{table*}[t]
    \centering
    \small{
    \caption{Ablation experiments of ClusVPR and pyramid self-supervised strategy (PSS).}
    \label{tab.ablation}
    \begin{tabular}{ l | c  c  c | c  c  c | c  c  c | c  c  c }
        \toprule[0.5pt]
        \multirow{2}{4em}{\textbf{Method}} & \multicolumn{3}{c|}{\textbf{MSLS val}} & \multicolumn{3}{c|}{\textbf{MSLS chall}} & \multicolumn{3}{c|}{\textbf{Pitts250k}} & \multicolumn{3}{c}{\textbf{Tokyo 24/7}} \\ \cmidrule(r){2-13}
        & R@1 & R@5 & R@10 & R@1 & R@5 & R@10 & R@1 & R@5 & R@10 & R@1 & R@5 & R@10 \\
        \midrule[0.5pt]
        NetVLAD + V-Triplet & 70.9 & 80.9 & 83.8 & 41.8 & 53.9 & 56.1 & 86.5 & 93.9 & 95.8 & 75.3 & 83.9 & 87.4 \\[-0.5ex]
	V-ClusVPR + V-Triplet & 74.2 & 81.3 & 84.6 & \textbf{44.6} & 55.8 & 56.9 & 88.4 & 95.4 & 96.4 & 79.9 & \textbf{87.1} & 90.5 \\[-0.5ex]
	ClusVPR + V-Triplet & \textbf{75.3} & \textbf{81.8} & \textbf{85.9} & 44.5 & \textbf{56.1} & \textbf{57.2} & \textbf{88.9} & \textbf{95.6} & \textbf{96.7} & \textbf{80.1} & 87.0 & \textbf{90.7} \\ 
        \midrule
	
	NetVLAD + S-Triplet & 76.4 & 81.7 & 85.1 & 44.6 & 55.9 & 56.7 & 89.1 & 95.6 & 96.8 & 80.4 & 87.3 & 90.8 \\[-0.5ex]
	V-ClusVPR + S-Triplet & 79.0 & 84.9 & \textbf{90.4} & 46.3 & 56.6 & 60.2 & 89.9 & \textbf{96.0} & 96.9 & 83.7 & \textbf{89.3} & 91.9 \\[-0.5ex]
	ClusVPR + S-Triplet & \textbf{79.8} & \textbf{85.1} & 90.3 & \textbf{47.2} & \textbf{56.9} & \textbf{60.4} & \textbf{90.2} & 95.9 & \textbf{97.0} & \textbf{84.6} & \textbf{89.9} & \textbf{92.2} \\
        \midrule
        
	NetVLAD + PSS & 77.4 & 82.9 & 85.8 & 46.1 & 56.4 & 58.2 & 90.5 & 96.2 & 97.1 & 85.1 & 90.3 & 92.9 \\[-0.5ex]
        V-ClusVPR + PSS & 81.6 & 87.8 & 91.6 & 48.9 & 58.1 & 62.0 & 91.7 & 96.6 & \textbf{97.6} & 86.7 & 91.1 & 93.6 \\[-0.5ex]
        ClusVPR + PSS & \textbf{82.7} & \textbf{88.5} & \textbf{92.4} & \textbf{49.4} & \textbf{58.2} & \textbf{62.3} & \textbf{92.4} & 96.9 & \textbf{97.6} & \textbf{87.5} & \textbf{91.5} & \textbf{93.9} \\
        \midrule
    \end{tabular}}
\end{table*}
The results of ablation experiments on the Pitts250k and Tokyo 24/7 are presented in Table~\ref{tab.ablation}. It is important to note that V-ClusVPR in our experiments refers to the ClusVPR with the standard NetVLAD instead of OptLAD. S-Triplet denotes softmax-based triplet loss, whereas V-Triplet denotes standard triplet loss. The following observations can be drawn from the experiments.

\noindent \textbf{Softmax-based triplet loss} shows superior efficacy compared to the standard triplet loss and can accelerate the convergence. Specifically, on the Pitts250k, the models with S-Triplet outperform the models with V-Triplet, with improvements on rank-1 recall of 2.6\%, 1.5\%, and 1.3\% respectively. 

\noindent \textbf{ClusVPR} successfully addresses the issue of unequal importance of image regions in VPR tasks and enhances the global reasoning ability of the model to encode geometric configurations between images. The results on the challenging Tokyo 24/7 show that the models with ClusVPR achieve significantly better rank-1 recall compared to models with NetVLAD, with improvements of 4.8\%, 4.2\%, and 2.4\% respectively, reaching 80.1\%, 84.6\%, and 87.5\%.

\noindent\textbf{OptLAD} improves the precision-recall performance of our model compared to those with NetVLAD, while having a significantly lower parameter count of around one-fourth. For example, the models with OptLAD outperform those with NetVLAD on the MSLS val, with improvements on rank-1 recall of 1.1\%, 0.8\%, and 1.1\% respectively. 

\noindent\textbf{Pyramid self-supervised strategy} has been shown to effectively improve performance by exploiting scale-wise image information. On the Tokyo 24/7, the models with PSS outperform those with S-Triplet, resulting in an improvement on rank-1 recall of 4.7\%, 3.0\%, and 2.9\%, respectively.

The impact of different down-sampling methods in the CWTNet is evaluated through an ablation experiment. Specifically, We compared the performance of our model with three different down-sampling methods (Average Pooling, Max Pooling, and Center Pooling), as shown in Table \ref{tab.downsample}. The results indicate that adopting the average pooling achieved the best precision on both the Pitts250k and Tokyo 24/7, with a marginal improvement of 0.6\% and 0.8\% on rank-1 recall compared to the second-best method, respectively. This implies that the average pooling is more effective in preserving the information from different regions of the image.

\begin{table}[t]
   \centering
   \small{
   \caption{Results of the ablation experiment on the down-sampling methods.}
   \label{tab.downsample}
   \setlength{\tabcolsep}{1.5mm}{
   \begin{tabular}{ l | c  c  c | c  c  c }
       \toprule[0.5pt]
       \multirow{2}{4em}{\textbf{Method}} & \multicolumn{3}{c|}{\textbf{Pitts250k}} & \multicolumn{3}{c}{\textbf{Tokyo 24/7}} \\ \cmidrule(r){2-7}
	& R@1 & R@5 & R@10 & R@1 & R@5 & R@10 \\
       \midrule[0.5pt]
       Average Pooling & \textbf{92.4} & \textbf{96.9} & \textbf{97.6} & \textbf{87.5} & \textbf{91.5} & \textbf{93.9} \\ [-0.5ex]
	Max Pooling & 91.8 & 96.6 & \textbf{97.6} & 86.7 & 90.9 & 93.3 \\ [-0.5ex]
	Center Pooling & 91.1 & 96.4 & 97.3 & 85.9 & 90.2 & 92.7 \\
       \bottomrule[0.5pt]
   \end{tabular}}}
\end{table}
\begin{figure*}[t]
\begin{center}
  \includegraphics[width=0.96\linewidth]{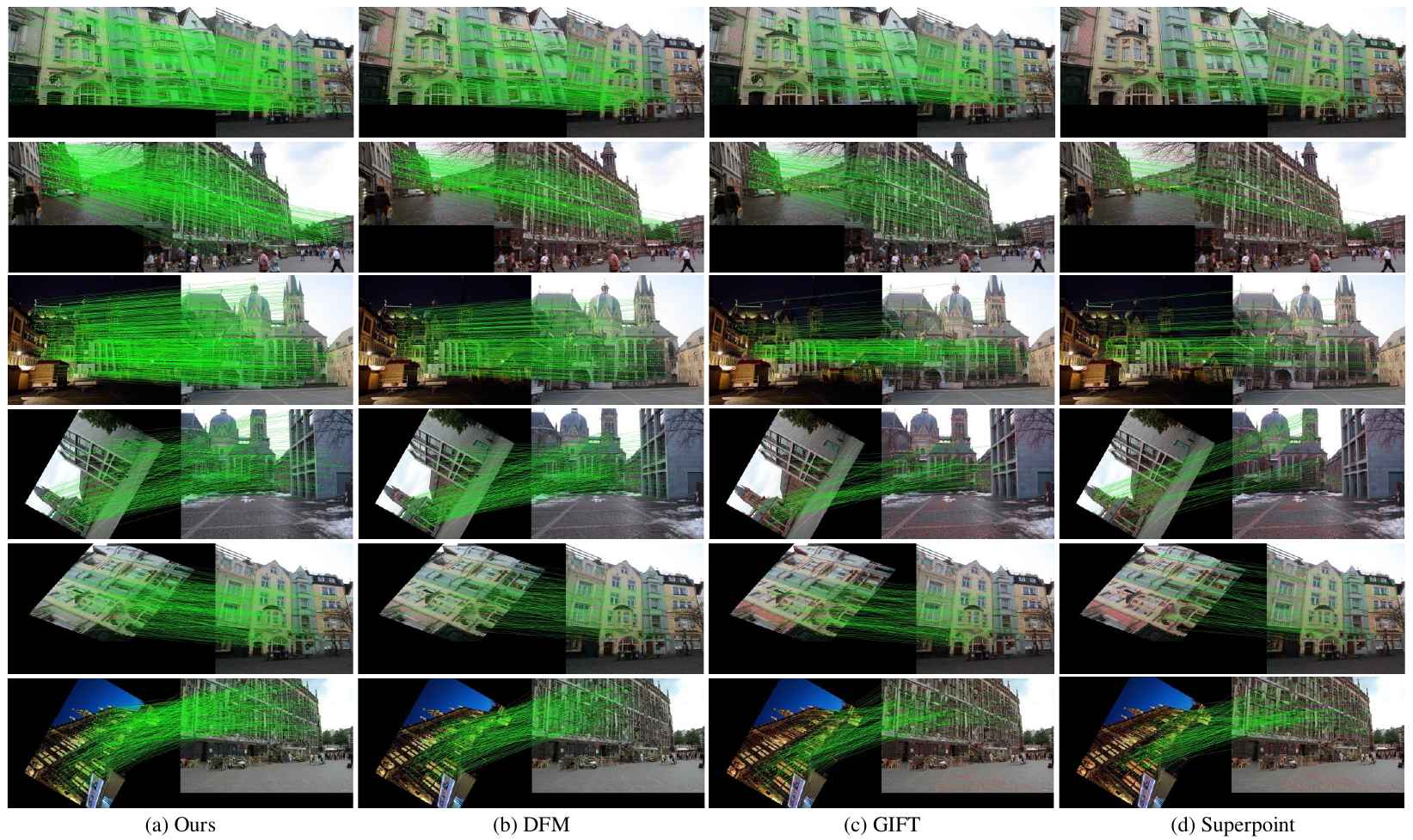}
\end{center}
  \caption{Qualitative matching results of different models. The models are evaluated on day, night, and rotated image pairs. Green lines indicate correct matches.}
\label{fig:keypoints}
\end{figure*}

\subsection{Keypoints detection and matching}
To further evaluate the effectiveness of our model, we conducted an experiment using the DFM \cite{Efe_2021_CVPRW} pipeline by replacing the VGG backbone with our pre-trained model. Since we only want to evaluate the performance of our pre-trained backbone in the DFM pipeline, we did not fine-tune it again. Instead, we simply replaced the feature extraction part with our pre-trained backbone and evaluated its performance on day, night, and rotated image pairs, as shown in Fig~\ref{fig:keypoints}. We compared the results with those of three strong baseline models-- DFM, GIFT \cite{Liu_2019_NIPS}, and Superpoint \cite{DeTone_2018_CVPR_Workshops}. The results demonstrate that our model produces more accurate and dense matches than the other three methods, indicating high generalization performance.

\subsection{Results with different CNN backbones}
To investigate whether using stronger backbones can improve the performance of our model on VPR datasets, we conducted additional experiments by employing VGG-19 \cite{Simonyan_2015_ICLR} and ResNets-101/152 \cite{He_2016_CVPR} as the backbones. We compare the results obtained with these backbones to our prior results on the VPR datasets.

\begin{figure}[t]
	\begin{center}
		\includegraphics[width=0.98\linewidth]{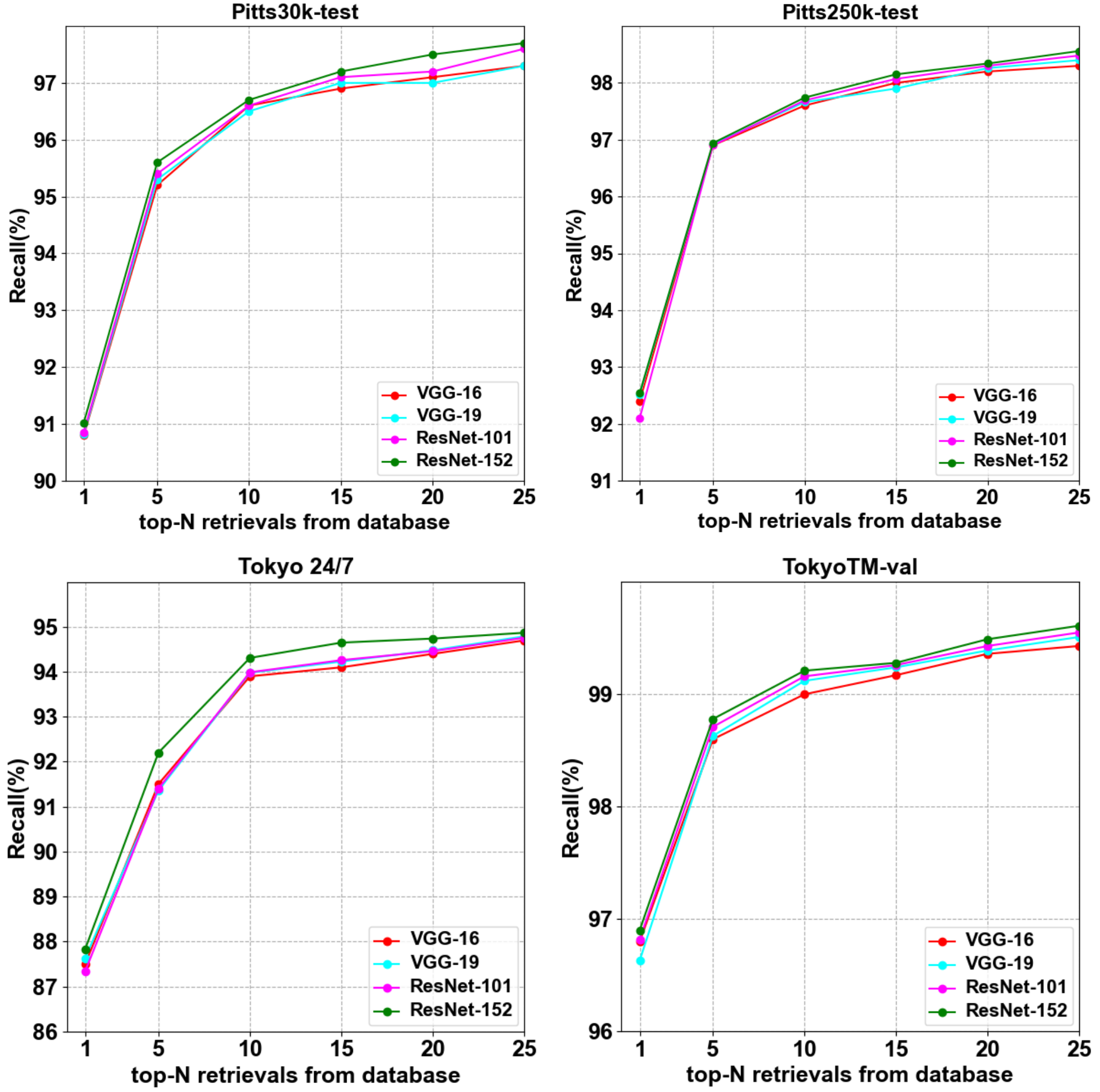}
	\end{center}
	\caption{Recalls at N-top retrievals for our ClusVPR with different CNN backbones.}
	\label{fig:PrecisionRecall_curve}
\end{figure}

The precision-recall curves of the models with different CNN backbones are displayed in Fig~\ref{fig:PrecisionRecall_curve}. Table~\ref{tab.different_CNNs} shows the model complexity and detailed precision-recall of the models. The results indicate that models with stronger backbones have less than 0.5\% rank-1 recall improvement compared to our ClusVPR with VGG-16. The advantages of utilizing stronger backbones are more evident in the Tokyo 24/7 dataset. For example, the model with ResNet-152 achieves 92.2\% rank-5 recall, a 0.7\% accuracy gain over ClusVPR with VGG-16, at the cost of 44.5M additional parameters. Consequently, after carefully evaluating the results of ClusVPR with different backbones, we conclude that using more robust and stronger backbones can yield marginal improvements in the model's performance, but at the cost of additional complexity.

\begin{table*}
	\centering
	\small{
		\caption{Model complexity and precision-recall of our ClusVPR with different CNN backbones on the Pitts250k and Tokyo 24/7 test sets.}
		\label{tab.different_CNNs}
		\setlength{\tabcolsep}{0.9mm}{
			\begin{tabular}{ l | c  c  c  c  c  c | c  c  c  c  c  c | c  c}
				\toprule[0.5pt]
				\multirow{2}{4em}{\textbf{Method}} & \multicolumn{6}{c|}{\textbf{Pitts250k}} & \multicolumn{6}{c|}{\textbf{Tokyo 24/7}} & \multicolumn{2}{c}{\textbf{Model Complexity}}\\ \cmidrule(r){2-15}
				
				& R@1 & R@5 & R@10 & R@15 & R@20 & R@25 & R@1 & R@5 & R@10 & R@15 & R@20 & R@25 & Params(M) & FLOPs(G) \\
				\midrule[0.5pt]
				
				ClusVPR-VGG-16 & 92.4 & \textbf{96.9} & 97.6 & 98.0 & 98.2 & 98.3 & 87.5 & 91.5 & 93.9 & 94.1 & 94.4 & 94.7 & \textbf{53.12} & 92.14 \\ [-0.3ex]
				
				
				ClusVPR-VGG-19 & \textbf{92.5} & \textbf{96.9} & \textbf{97.7} & 97.9 & \textbf{98.3} & 98.4 & 87.6 & 91.4 & 94.0 & 94.2 & 94.5 & 94.8 & 58.46 & 121.97 \\[-0.3ex]
				
				ClusVPR-ResNet-101 & 92.1 & \textbf{96.9} & \textbf{97.7} & 98.1 & \textbf{98.3} & 98.5 & 87.3 & 91.4 & 94.0 & 94.3 & 94.5 & 94.8 & 81.98 & \textbf{50.61} \\[-0.3ex]
				
				ClusVPR-ResNet-152 & \textbf{92.5} & \textbf{96.9} & \textbf{97.7} & \textbf{98.2} & \textbf{98.3} & \textbf{98.6} & \textbf{87.8} & \textbf{92.2} & \textbf{94.3} & \textbf{94.7} & \textbf{94.7} & \textbf{94.9} & 97.62 & 73.49 \\
				\bottomrule[0.5pt]
	\end{tabular}}}
\end{table*}

\section{Conclusions}
\label{sec:conclusions}
This paper proposed a novel ClusVPR model and CWTNet architecture that improves the global representation of images for VPR tasks, addressing the problem of the unequal importance of image regions. The CWTNet calculates weights for image tokens and encodes global dependencies, allowing our model to encode corresponding geometric configurations between images. The proposed OptLAD layer enhances the efficiency of the model and outperforms the standard NetVLAD. Additionally, a pyramid self-supervised strategy is introduced to extract more precise information from scale-wise image patches. The extensive experiments demonstrate the superior performance of our proposed model in terms of qualitative, quantitative, and efficiency evaluations.


 
 \bibliographystyle{IEEEtran} 
 \bibliography{cas-refs}

\begin{thebibliography}{10}
\providecommand{\url}[1]{#1}
\csname url@samestyle\endcsname
\providecommand{\newblock}{\relax}
\providecommand{\bibinfo}[2]{#2}
\providecommand{\BIBentrySTDinterwordspacing}{\spaceskip=0pt\relax}
\providecommand{\BIBentryALTinterwordstretchfactor}{4}
\providecommand{\BIBentryALTinterwordspacing}{\spaceskip=\fontdimen2\font plus
\BIBentryALTinterwordstretchfactor\fontdimen3\font minus
  \fontdimen4\font\relax}
\providecommand{\BIBforeignlanguage}[2]{{%
\expandafter\ifx\csname l@#1\endcsname\relax
\typeout{** WARNING: IEEEtran.bst: No hyphenation pattern has been}%
\typeout{** loaded for the language `#1'. Using the pattern for}%
\typeout{** the default language instead.}%
\else
\language=\csname l@#1\endcsname
\fi
#2}}
\providecommand{\BIBdecl}{\relax}
\BIBdecl

\bibitem{arandjelovic_2018_netvlad}
R.~Arandjelović, P.~Gronat, A.~Torii, T.~Pajdla, and J.~Sivic, ``Netvlad: Cnn
  architecture for weakly supervised place recognition,'' \emph{IEEE Trans.
  Pattern Anal. Mach. Intell.}, vol.~40, no.~6, pp. 1437--1451, 2018.

\bibitem{Fan_2022_TMM}
B.~Fan, Y.~Yang, W.~Feng, F.~Wu, J.~Lu, and H.~Liu, ``Seeing through darkness:
  Visual localization at night via weakly supervised learning of domain
  invariant features,'' \emph{IEEE Trans. Multimedia}, 2022.

\bibitem{Khaliq_2020_Robotics}
A.~Khaliq, S.~Ehsan, Z.~Chen, M.~Milford, and K.~McDonald-Maier, ``A holistic
  visual place recognition approach using lightweight cnns for significant
  viewpoint and appearance changes,'' \emph{IEEE Trans. Robot.}, 2020.

\bibitem{HANE_2017_ICV}
C.~Häne, L.~Heng, G.~H. Lee, F.~Fraundorfer, P.~Furgale, T.~Sattler, and
  M.~Pollefeys, ``3d visual perception for self-driving cars using a
  multi-camera system: Calibration, mapping, localization, and obstacle
  detection,'' \emph{Image Vis. Comput.}, 2017.

\bibitem{He_2016_CVPR}
K.~He, X.~Zhang, S.~Ren, and J.~Sun, ``Deep residual learning for image
  recognition,'' in \emph{IEEE Conf. Comput. Vis. Pattern Recog.}, 2016.

\bibitem{Hsu_2018_TMM}
C.-C. Hsu and C.-W. Lin, ``Cnn-based joint clustering and representation
  learning with feature drift compensation for large-scale image data,''
  \emph{IEEE Trans. Multimedia}, 2018.

\bibitem{Alexander_2021_ICLR}
A.~Dosovitskiy, L.~Beyer, A.~Kolesnikov, and D.~Weissenborn, ``An image is
  worth 16x16 words: Transformers for image recognition at scale,'' in
  \emph{Int. Conf. Learn. Represent.}, 2021.

\bibitem{Lin_2021_TMM}
X.~Lin, S.~Sun, W.~Huang, B.~Sheng, P.~Li, and D.~D. Feng, ``Eapt: Efficient
  attention pyramid transformer for image processing,'' \emph{IEEE Trans.
  Multimedia}, 2021.

\bibitem{Ge_2020_ECCV}
Y.~Ge, H.~Wang, F.~Zhu, R.~Zhao, and H.~Li, ``Self-supervising fine-grained
  region similarities for large-scale image localization,'' in \emph{Eur. Conf.
  Comput. Vis.}, 2020.

\bibitem{yang_2021_NIPS}
y.~hongji, X.~Lu, ., and Y.~Zhu, ``Cross-view geo-localization with
  layer-to-layer transformer,'' in \emph{Adv. Neural Inform. Process. Syst.},
  2021.

\bibitem{Hausler_2021_CVPR}
S.~Hausler, S.~Garg, M.~Xu, M.~Milford, and T.~Fischer, ``Patch-netvlad:
  Multi-scale fusion of locally-global descriptors for place recognition,'' in
  \emph{IEEE Conf. Comput. Vis. Pattern Recog.}, 2021.

\bibitem{leyva2023data}
M.~Leyva-Vallina, N.~Strisciuglio, ., and N.~Petkov, ``Data-efficient large
  scale place recognition with graded similarity supervision,'' in \emph{IEEE
  Conf. Comput. Vis. Pattern Recog.}, 2023.

\bibitem{Arandjelovic_2016_CVPR}
R.~Arandjelović, P.~Gronat, A.~Torii, T.~Pajdla, and J.~Sivic, ``Netvlad: Cnn
  architecture for weakly supervised place recognition,'' in \emph{IEEE Conf.
  Comput. Vis. Pattern Recog.}, 2016.

\bibitem{ali2023global}
A.~Ali-bey, Chaib-draa, Brahim, and P.~Gigu{\`e}re, ``Global proxy-based hard
  mining for visual place recognition,'' \emph{British Mach. Vis. Conf.}, 2023.

\bibitem{Alaaeldin_2021_arXiv}
A.~El-Nouby, N.~Neverova, I.~Laptev, and H.~Jégou, ``Training vision
  transformers for image retrieval,'' \emph{arXiv preprint arXiv:2102.05644},
  2021.

\bibitem{Zhu_2023_arXiv}
S.~Zhu, L.~Yang, C.~Chen, M.~Shah, X.~Shen, and H.~Wang, ``$r^{2}$former:
  Unified $r$etrieval and $r$eranking transformer for place recognition,''
  \emph{arXiv preprint arXiv:2304.03410}, 2023.

\bibitem{Child_2019_arXiv}
R.~Child, S.~Gray, A.~Radford, and I.~Sutskever, ``Generating long sequences
  with sparse transformers,'' \emph{arXiv preprint arXiv:1904.10509}, 2019.

\bibitem{Liu_2019_ICCV}
L.~Liu, H.~Li, and Y.~Dai, ``Stochastic attraction-repulsion embedding for
  large scale image localization,'' in \emph{Int. Conf. Comput. Vis.}, 2019,
  pp. 2570--2579.

\bibitem{Schroff_2015_CVPR}
F.~Schroff, D.~Kalenichenko, ., and J.~Philbin, ``Facenet: A unified embedding
  for face recognition and clustering,'' in \emph{IEEE Conf. Comput. Vis.
  Pattern Recog.}, 2015.

\bibitem{Cummins_2008_IJRR}
M.~Cummins and P.~Newman, ``Fab-map: Probabilistic localization and mapping in
  the space of appearance,'' \emph{Int. J. Rob. Res.}, 2008.

\bibitem{Mur_2017_TR}
R.~Mur-Artal and J.~D. Tard{\'o}s, ``Orb-slam2: An open-source slam system for
  monocular, stereo, and rgb-d cameras,'' \emph{IEEE Trans. Robot.}, 2017.

\bibitem{Lowe_1999_ICCV}
D.~G. Lowe, ``Object recognition from local scale-invariant features,'' in
  \emph{Int. Conf. Comput. Vis.}, 1999.

\bibitem{Bay_2006_ECCV}
H.~Bay, T.~Tuytelaars, and L.~V. Gool, ``Surf: Speeded up robust features,'' in
  \emph{Eur. Conf. Comput. Vis.}, 2006.

\bibitem{Xu_2023_WACV}
Y.~Xu, P.~Shamsolmoali, and E.~Granger, ``Transvlad: Multi-scale
  attention-based global descriptors for visual geo-localization,'' in
  \emph{IEEE Winter Conf. Appl. Comput. Vis.}, 2023.

\bibitem{cao2020unifying}
B.~Cao, A.~Araujo, and J.~Sim, ``Unifying deep local and global features for
  image search,'' in \emph{Eur. Conf. Comput. Vis.}, 2020.

\bibitem{zhu2018attention}
Y.~Zhu, J.~Wang, L.~Xie, and L.~Zheng, ``Attention-based pyramid aggregation
  network for visual place recognition,'' in \emph{Pro. ACM Int. Conf.
  Multimedia}, 2018.

\bibitem{radenovic2018fine}
F.~Radenovi{\'c}, G.~Tolias, ., and O.~Chum, ``Fine-tuning cnn image retrieval
  with no human annotation,'' \emph{IEEE Trans. Pattern Anal. Mach. Intell.},
  vol.~41, no.~7, 2018.

\bibitem{gordo2017end}
A.~Gordo, J.~Almazan, J.~Revaud, and D.~Larlus, ``End-to-end learning of deep
  visual representations for image retrieval,'' \emph{Int. J. Comput. Vis.},
  vol. 124, no.~2, 2017.

\bibitem{berton2022deep}
G.~Berton, R.~Mereu, G.~Trivigno, C.~Masone, G.~Csurka, T.~Sattler, and
  B.~Caputo, ``Deep visual geo-localization benchmark,'' in \emph{IEEE Conf.
  Comput. Vis. Pattern Recog.}, 2022.

\bibitem{shen2023structvpr}
Y.~Shen, S.~Zhou, J.~Fu, R.~Wang, S.~Chen, and N.~Zheng, ``Structvpr: Distill
  structural knowledge with weighting samples for visual place recognition,''
  in \emph{IEEE Conf. Comput. Vis. Pattern Recog.}, 2023.

\bibitem{Vaswani_2017_NIPS}
A.~Vaswani, N.~Shazeer, N.~Parmar, J.~Uszkoreit, L.~Jones, A.~N. Gomez, L.~u.
  Kaiser, and I.~Polosukhin, ``Attention is all you need,'' in \emph{Adv.
  Neural Inform. Process. Syst.}, 2017.

\bibitem{Touvron_2021_ICML}
H.~Touvron, M.~Cord, M.~Douze, F.~Massa, A.~Sablayrolles, and H.~Jegou,
  ``Training data-efficient image transformers \& amp; distillation through
  attention,'' in \emph{Proc. Int. Conf. Mach. Learn.}, 2021.

\bibitem{Wang_2022_CVPR}
R.~Wang, Y.~Shen, W.~Zuo, S.~Zhou, and N.~Zheng, ``Transvpr: Transformer-based
  place recognition with multi-level attention aggregation,'' in \emph{IEEE
  Conf. Comput. Vis. Pattern Recog.}, 2022.

\bibitem{Yan_2021_ICASSP}
L.~Yan, Y.~Cui, Y.~Chen, and D.~Liu, ``Hierarchical attention fusion for
  geo-localization,'' in \emph{IEEE Int. Conf. Acoust. Speech. Signal. Process.
  Proc.}, 2021.

\bibitem{hu2020dasgil}
H.~Hu, Z.~Qiao, M.~Cheng, Z.~Liu, and H.~Wang, ``Dasgil: Domain adaptation for
  semantic and geometric-aware image-based localization,'' \emph{IEEE Trans.
  Image Process.}, vol.~30, 2020.

\bibitem{paolicelli2022learning}
V.~Paolicelli, A.~Tavera, C.~Masone, G.~Berton, and B.~Caputo, ``Learning
  semantics for visual place recognition through multi-scale attention,'' in
  \emph{Int. Conf. Image Anal. Process.}, 2022.

\bibitem{wu_2021_ICCV}
H.~Wu, B.~Xiao, N.~Codella, M.~Liu, X.~Dai, L.~Yuan, and L.~Zhang, ``Cvt:
  Introducing convolutions to vision transformers,'' in \emph{Int. Conf.
  Comput. Vis.}, 2021.

\bibitem{Hendrycks_2016_arXiv}
D.~Hendrycks and K.~Gimpel, ``Gaussian error linear units (gelus),''
  \emph{arXiv preprint arXiv:1606.08415}, 2016.

\bibitem{Herv_2012_ECCV}
H.~J{\'e}gou and O.~Chum, ``Negative evidences and co-occurences in image
  retrieval: The benefit of pca and whitening,'' in \emph{Eur. Conf. Comput.
  Vis.}, 2012.

\bibitem{Arandjelovic_2013_CVPR}
R.~Arandjelović and A.~Zisserman, ``All about vlad,'' in \emph{IEEE Conf.
  Comput. Vis. Pattern Recog.}, 2013.

\bibitem{Torii_2015_CVPR}
A.~Torii, R.~Arandjelović, J.~Sivic, M.~Okutomi, and T.~Pajdla, ``24/7 place
  recognition by view synthesis,'' in \emph{IEEE Conf. Comput. Vis. Pattern
  Recog.}, 2015.

\bibitem{warburg2020mapillary}
F.~Warburg, S.~Hauberg, M.~Lopez-Antequera, P.~Gargallo, Y.~Kuang, and
  J.~Civera, ``Mapillary street-level sequences: A dataset for lifelong place
  recognition,'' in \emph{IEEE Conf. Comput. Vis. Pattern Recog.}, 2020.

\bibitem{Torii_2013_CVPR}
A.~Torii, J.~Sivic, T.~Pajdla, and M.~Okutomi, ``Visual place recognition with
  repetitive structures,'' in \emph{IEEE Conf. Comput. Vis. Pattern Recog.},
  2013.

\bibitem{Philbin_2008_CVPR}
J.~Philbin, O.~Chum, ., and M.~Isard, ``Lost in quantization: Improving
  particular object retrieval in large scale image databases,'' in \emph{IEEE
  Conf. Comput. Vis. Pattern Recog.}, 2008.

\bibitem{Philbin_2007_CVPR}
J.~Philbin, O.~Chum, M.~Isard, J.~Sivic, and A.~Zisserman, ``Object retrieval
  with large vocabularies and fast spatial matching,'' in \emph{IEEE Conf.
  Comput. Vis. Pattern Recog.}, 2007.

\bibitem{Jegou_2008_ECCV}
H.~Jegou, M.~Douze, and C.~Schmid, ``Hamming embedding and weak geometric
  consistency for large scale image search,'' in \emph{Eur. Conf. Comput.
  Vis.}, 2008.

\bibitem{Stylianou_2019_WACV}
A.~Stylianou, R.~Souvenir, ., and R.~Pless, ``Visualizing deep similarity
  networks,'' in \emph{IEEE Winter Conf. Appl. Comput. Vis.}, 2019.

\bibitem{Efe_2021_CVPRW}
U.~Efe, K.~G. Ince, and A.~Alatan, ``Dfm: A performance baseline for deep
  feature matching,'' in \emph{IEEE Conf. Comput. Vis. Pattern Recog.
  Workshops}, 2021.

\bibitem{Liu_2019_NIPS}
Y.~Liu, Z.~Shen, Z.~Lin, S.~Peng, H.~Bao, and X.~Zhou, ``Gift: Learning
  transformation-invariant dense visual descriptors via group cnns,'' in
  \emph{Adv. Neural Inform. Process. Syst.}, 2019.

\bibitem{DeTone_2018_CVPR_Workshops}
D.~DeTone, T.~Malisiewicz, and A.~Rabinovich, ``Superpoint: Self-supervised
  interest point detection and description,'' in \emph{IEEE Conf. Comput. Vis.
  Pattern Recog. Workshops}, 2018.

\bibitem{Simonyan_2015_ICLR}
K.~Simonyan and A.~Zisserman, ``Very deep convolutional networks for
  large-scale image recognition,'' in \emph{Int. Conf. Learn. Represent.},
  2015.

\end{thebibliography}
%

\newpage

 




\vfill

\end{document}